\documentclass[10pt, conference]{IEEEtran}
\pdfoutput=1
\usepackage{cite}
\usepackage{amsmath,amssymb,amsfonts}

\usepackage{cite}
\usepackage{dsfont}
\usepackage{balance}
\usepackage{diagbox}
\usepackage{algorithmic}
\usepackage{algorithm}
\usepackage{graphicx}
\graphicspath{ {./images/} }
\usepackage{textcomp}
\usepackage[dvipsnames]{xcolor}

\makeatletter

\let\old@ps@headings\ps@headings
\let\old@ps@IEEEtitlepagestyle\ps@IEEEtitlepagestyle
\def\confheader#1{%
    \def\ps@IEEEtitlepagestyle{%
        \old@ps@IEEEtitlepagestyle%
        \def\@oddhead{\strut\hfill#1\hfill\strut}%
        \def\@evenhead{\strut\hfill#1\hfill\strut}%
    }%
    \ps@headings%
}
\makeatother

\confheader{%
        \parbox{\textwidth}{\centering This article has been accepted for publication in the IEEE International Conference on Sensing, Communication, and Networking (SECON Demo), 2022.}
}

\usepackage[pscoord]{eso-pic}
\newcommand{\placetextbox}[3]{
\setbox0=\hbox{#3}
\AddToShipoutPictureFG{ \put(\LenToUnit{#1\paperwidth},\LenToUnit{#2\paperheight}){\vtop{{\null}\makebox[0pt][c]{#3}}}
}
}
\placetextbox{.5}{0.065}{\footnotesize {\copyright 2022 IEEE. Personal use of this material is permitted. Permission from IEEE must be obtained for all other uses, in any current or future media} }
\placetextbox{.5}{0.055}{\footnotesize { including reprinting/republishing this material for advertising or promotional purposes, creating new  collective works, for resale}} 
\placetextbox{.5}{0.045}{\footnotesize { or redistribution to servers or lists, or reuse of any copyrighted component of this work in other works.}
}

\begin{document}

\title{Edge-assisted Collaborative Digital Twin for Safety-Critical Robotics in Industrial IoT}

\author{\IEEEauthorblockN{
\large Sumit K. Das\IEEEauthorrefmark{1}, Mohammad Helal Uddin\IEEEauthorrefmark{2}, Sabur Baidya\IEEEauthorrefmark{2}\\
\IEEEauthorblockA{\IEEEauthorrefmark{1}\normalsize Department of Electrical and Computer Engineering, University of Louisville, KY, USA}
\IEEEauthorblockA{\IEEEauthorrefmark{2}\normalsize Department of Computer Science and Engineering, University of Louisville, KY, USA}
\normalsize {e-mail:   sumitkumar.das@louisville.edu, mohammad.helaluddin@louisville.edu, sabur.baidya@louisville.edu}
%\vspace{-6mm}
}}

\maketitle

\vspace{-4mm}
\begin{abstract}
Digital Twin technology is playing a pivotal role in the modern industrial evolution. Especially, with the technological progress in the Internet-of-Things (IoT) and the increasing trend in autonomy, multi-sensor equipped robotics can create practical digital twin, which is particularly useful in the industrial applications for operations, maintenance and safety. Herein, we demonstrate a real-world digital twin of a safety-critical robotics applications with a Franka-Emika-Panda robotic arm. We develop and showcase an edge-assisted collaborative digital twin for  dynamic obstacle avoidance which can be useful in real-time adaptation of the robots while operating in the uncertain and dynamic environments in industrial IoT.

\end{abstract}

%\begin{IEEEkeywords}
 %Digital Twin, Robotics, Internet of Things, Industrial IoT, Robotic Arm, Robot Operating System (ROS) 
%\end{IEEEkeywords}

\vspace{-3mm}
\section{Introduction}
The emergence of the Industry 4.0 has revolutionized the modern industrial applications. Especially, the rapid advancement in the automation and the Internet-of-Things (IoT) technologies~\cite{li2015internet}, have enabled increasing applications of robotics in the industrial domains for autonomous operations, enhanced scalability and reducing cost. Digital twin, on the other hand, is a framework where a physical entity and its surrounding environment can be reproduced digitally for remote operations, maintenance and safety. In the field of robotics, the digital twin technology is revolutionizing many industrial applications including the manufacturing, healthcare, connected and autonomous vehicles, construction and many others~\cite{qi2019enabling}. Herein, we address a scenario of safety-critical industrial robotics application, where an autonomous operation of a robotic arm can come across uncertainties or dynamics in the environment. A complete autonomous adaptation of the robot coping with the uncertainties~\cite{matulis2021robot} may not be safety-proof and can cause multifold damage in the robotic operations and cost. To mitigate the problem, a robust digital-twin can help remotely monitoring the behavior of the robot and the environment in real-time, creating a blocking closed-loop operation ensuring the safety. We propose the digital twin will be collaborative and edge-assisted, i.e., autonomously operated until uncertainties are detected by an edge computing unit, upon which the digital twin will re-plan the robot motions and deploy when it is safe. Note that the industrial robotics applications are safety-critical but not mission-critical. So, a blocking closed-loop does not  fail the mission but increases safety. Additionally, unlike an offline digital twin where problem is resolved offline after long time, or a complete teleoperation which is expensive in terms of human effort and cost, our edge-assisted collaborative digital twin ensures safety while minimizing the cost of resolving the uncertainties in operation.

\begin{figure}[!t]
\centering
\includegraphics[width=0.95\linewidth]{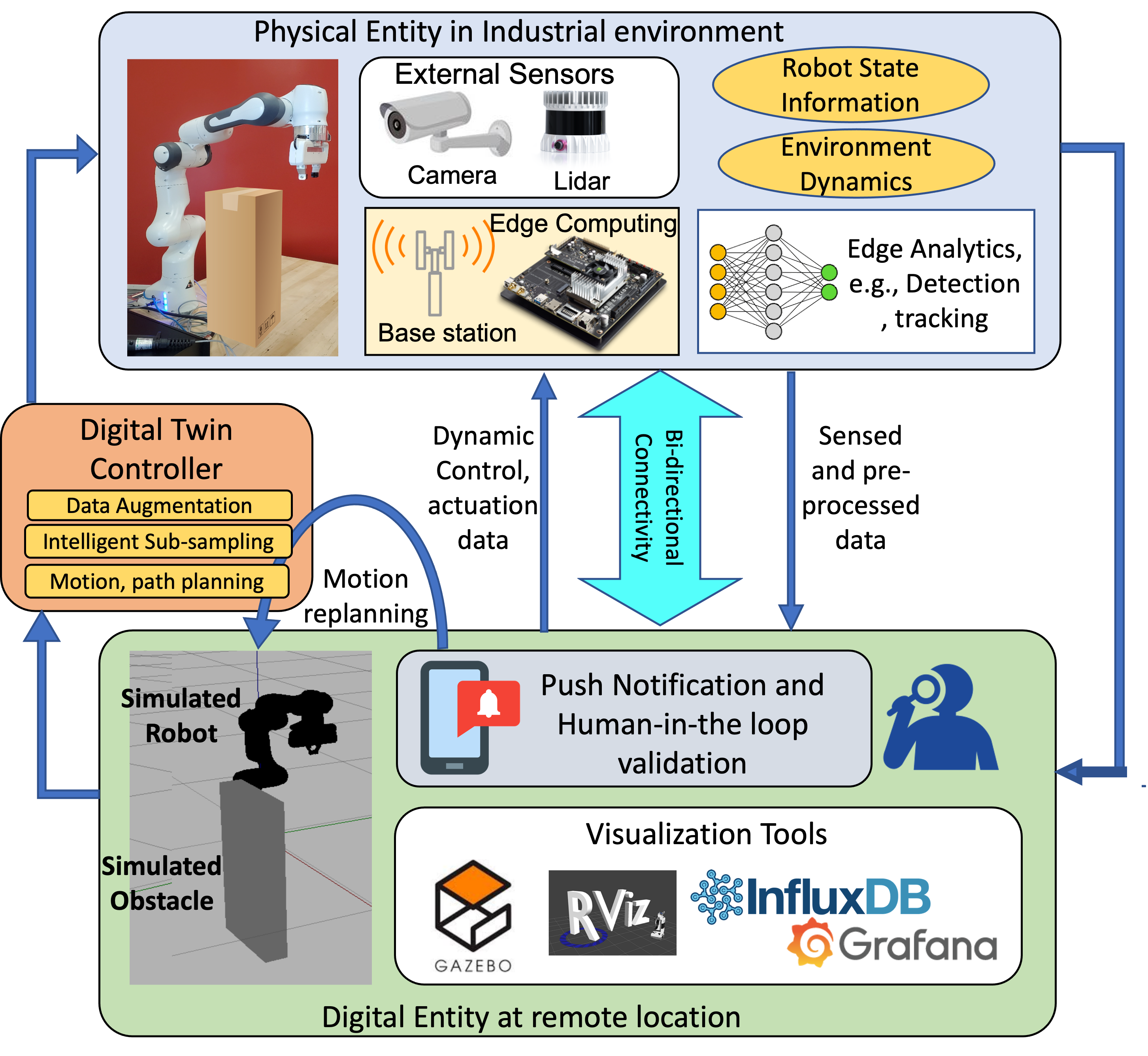}
\vspace{-2mm}
\caption{Edge-assisted Human-in-the-loop Architecture of Robotic  Digital Twin}
\label {fig:twin_acrh}
\vspace{-4mm}
\end{figure}

\section{System Overview}
Fig.~\ref{fig:twin_acrh} shows the overall architecture of our edge-assisted collaborative digital twin of a safety-critical robotic arm application. It consists of the following components:

\noindent \underline{\it Physical Entity:} The physical entity consists of the real-world robotic arm and associated sensors, actuators and other physical devices in the surrounding environment. It also contains a physical process or operation running with the robot.

\noindent \underline{\it Digital Entity:}  The digital entity, usually created in a remote location, consists of the digital replica of the robot along with its ambience is reproduced. It also reproduces the physics, kinematics of the behavior of the robot in simulation.

\noindent \underline{\it Edge Computing:} The edge computing unit resides at the site of the physical robot in the industry and help processing the information acquired by the IoT sensors. The edge computing helps in running AI-driven algorithms for detection of uncertainties in the environment, and the preprocessed information can be conveyed to its digital twin with much lower latency than sharing raw data. 

\noindent \underline{\it Digital Twin Controller:} The edge informed dynamic information is replicated in soft-real-time on the digital entity at the digital twin controller which evaluates if any changes in motion plan is needed. If needed, it triggers autonomous algorithms for adaptation at the digital entity first, then evaluate by a human in the loop for validation before deploying updated motion plan to the physical counterpart.

\noindent \underline{\it Visualization Tools:} The digital site also consists of extensive visualization tools that not only shows the robotic movement in real-time but also reproduces any dynamic changes in the environment, e.g., an obstacle in digital simulation. Additionally, the performance of the physical vs digital robots including movements, speed, states can be monitored in real-time plots with advanced IoT visualization tools.

\begin{figure}[!t]
\centering
\includegraphics[width=0.85\linewidth]{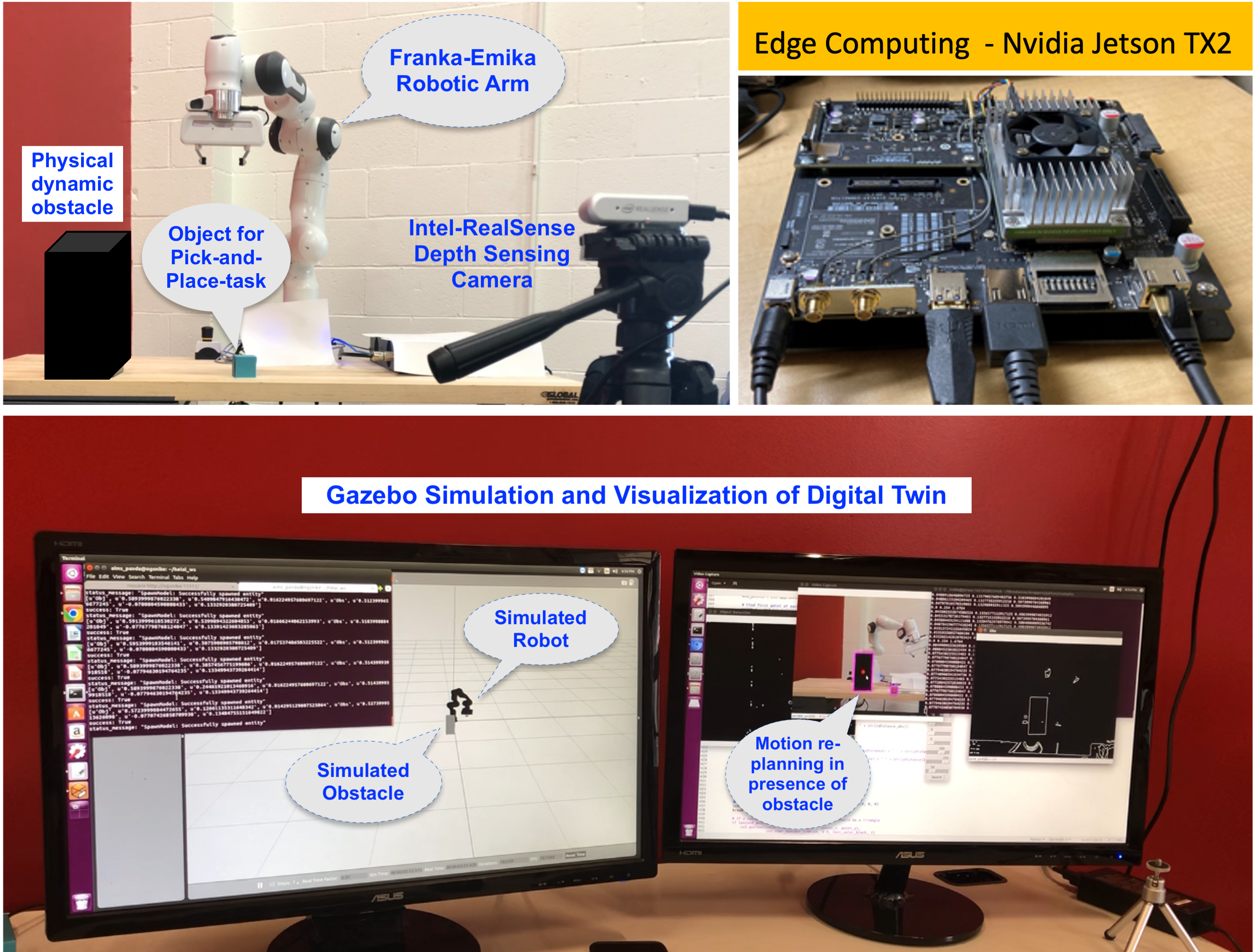}
\vspace{-2mm}
\caption{Demo Setup for Franka Robotic Arm Digital Twin}
\label {fig:twin_setup}
\vspace{-5mm}
\end{figure}

\vspace{-2mm}
\section{Setup, Implementation and Results}

We use a Franka-Emika-Panda robotic arm which can be used for various pick and place tasks in industry floor and can be operated with Robot Operating System (ROS). For digital counterpart, we use Gazebo simulation that also integrates with ROS~\cite{qian2014manipulation}. We employ a pick and place task of a small object with a gripper end-effector on the robotic arm. The main study we conduct over this framework is a dynamic obstacle avoidance as shown in fig.~\ref{fig:twin_setup}. We dynamically place a larger box as an obstacle in the experiment that can be placed at any location within the trajectory of the robot while performing the task. For detection of the uncertainty, we use an external Intel RealSense depth sensing camera that can detect the obstacle based on size, color or other feature. The detection is done at the edge computing node, Nvidia Jetson TX2~\cite{Nvidia_JTX}, connected to the camera at the robot site. The edge computing accelerates the detection, thus stops the robots on any anomaly due to introduced uncertainty to provide enhanced safety. Edge computing also reduces the data volume to be transmitted over the network to the remote location, thus the communication latency of the digital twin.

The digital counterpart receives the  size and coordinates of the location of the dynamic obstacle and renders the simulated obstacle in the digital robot environment. This triggers an automated updated motion plan and also a message to a human in the loop for validation. Once the new motion plan is validated, the digital-twin controller deploys it to the physical robot which safely and effectively avoids the obstacle.

\begin{figure}[!t]
\centering
\includegraphics[width=0.90\linewidth]{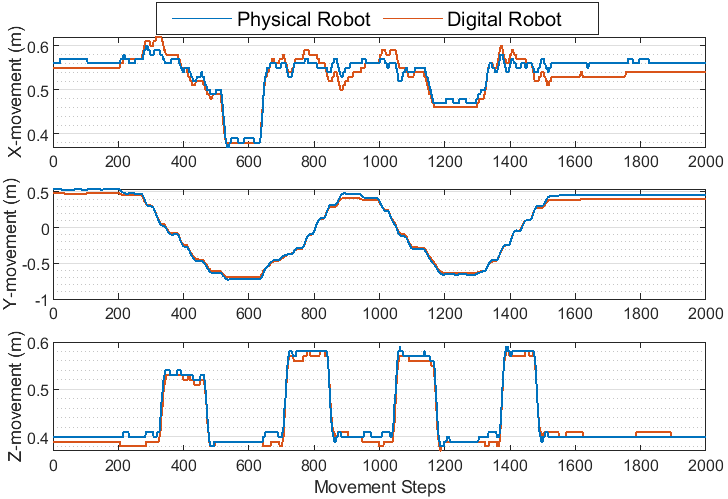}
\vspace{-3mm}
%\caption{Temporal translational deviation in motion between a physical robot and digital twin robot}
\caption{Temporal translational deviation  between a physical and digital robot}
\label{fig:dt_trans_err}
\vspace{-1mm}
\end{figure}

Fig.~\ref{fig:dt_trans_err} shows the translational movements of both physical and digital robot and a very small differences can be noticed in the order of few centimeters. The mean absolute error (MAE) for the movements which shows the errors less than 3 cm as mentioned in the table I. 

We also measure the latency of edge computing, communication and actuation delay between physical and digital robot in table I. The edge computing delay is somewhat large due to image segmentation, detection and localization, and will give us the opportunity to optimize in future. 

%the mean absolute difference in actuation time is about 27 ms, which can be considered as real-time. 
%The slight error could be created due to the difference in controller gain between the physical and digital robot. One can set up the safety margin slightly greater than this errors for seamless operation. 

\begin{table}[!t]
\centering
\begin{tabular}{|c|c|c|c|c|c|} 
 \hline
 
 {\bf X-mov} & {\bf Y-mov} & {\bf Z-mov} & {\bf Edge} & {\bf Commun-} & {\bf Actuation}  \\ [0.5ex] 
 {\bf Error} & {\bf Error} & {\bf Error}  & {\bf Comp.} & {\bf -ication} & {\bf Latency}\\ [0.5ex] 
  {\bf (m)} & {\bf (m)} & {\bf (m)}  & {\bf Time (ms)} & {\bf Delay (ms)} & {\bf Time (ms)}\\ [0.5ex] 
 
 \hline
 0.016 & 0.03 & 0.008 & 375 & 28.79 & 27.35 \\ [0.5ex] 
 \hline
\end{tabular}
\vspace{1mm}
\caption{\small MAE in translational movements \& different latencies}
\label{Accu_vs_Compl}
\vspace{-9mm}
\end{table}

%In this implementation of the digital twin, we successfully created a ROS based bidirectional control for the robotic arm. The performance from these preliminary experiments show promise of successful development of our proposed safety-proof human-in-the-loop framework with enhanced algorithms in future. 

\vspace{-1mm}
\section{Conclusion}
In this paper, we demonstrated an edge-assisted collaborative digital twin of a robotic arm and presented the performance in dynamic uncertain conditions. We proposed a human-in-the-loop architecture for ensuring the safety of the robots during dynamic adaptation. In future, the framework can be augmented with more sophisticated detection and tracking algorithm and innovation in edge computing and wireless communication to make it more efficient and robust.

\vspace{-1mm}
\section*{Acknowledgment}
This work is partially supported by the US National Science Foundation (EPSCoR \#1849213).

\bibliographystyle{IEEEtran}
\bibliography{digital-twin-bib}

\end{document}